\documentclass[11pt,a4paper]{article}
\usepackage[hyperref]{acl2017}
\usepackage{times}
\usepackage{latexsym}
\usepackage[pdftex]{graphicx}
\usepackage{comment}

\usepackage{url}

\aclfinalcopy 

\title{Word-Alignment-Based Segment-Level Machine Translation Evaluation using Word Embeddings}

\author{Junki Matsuo \and Mamoru Komachi\\
  Graduate School of System Design,\\
  Tokyo Metropolitan University, Japan \\
  {\tt matsuo-junki@ed.tmu.ac.jp,} \\
  {\tt komachi@tmu.ac.jp} \\\And
  Katsuhito Sudoh \\
  NTT Communication Science \\
  Laboratories, Japan \\
  {\tt sudoh@is.naist.jp}\thanks{The last author is
  currently affiliated with Nara Institute of Science and Technology, Japan.} \
  \\}

\date{}

\begin{document}
\maketitle
\begin{abstract}
One of the most important problems in machine translation (MT) evaluation is to evaluate the similarity between translation hypotheses with different surface forms from the reference, especially at the segment level. We propose to use word embeddings to perform word alignment for segment-level MT evaluation. We performed experiments with three types of alignment methods using word embeddings. We evaluated our proposed methods with various translation datasets. Experimental results show that our proposed methods outperform previous word embeddings-based methods.
\end{abstract}

\section{Introduction}
Automatic evaluation of machine translation (MT) systems without human intervention has gained importance. For example, BLEU \cite{Papineni:2002} has improved the MT research in the last decade. However, BLEU has little correlation with human judgment on the segment level since it is originally proposed for system-level evaluation. Segment-level evaluation is crucial for analyzing MT outputs to improve the system accuracy, but there are few studies addressing the issue of segment-level evaluation of MT outputs. 

Another issue in MT evaluation is to evaluate MT hypotheses that are semantically equivalent with different surfaces from the reference. For instance, BLEU does not consider any words that do not match the reference at the surface level. METEOR-Universal \cite{Denkowski:2014} handles word similarities better, but it uses external resources that require time-consuming annotations. It is also not as simple as BLEU and its score is difficult to interpret. DREEM \cite{chen-guo:2015:ACL-IJCNLP}, another metric that addresses the issue of word similarity, does not require human annotations and uses distributed representations for MT evaluation. It shows higher accuracy than popular metrics such as BLEU and METEOR.

Therefore, we follow the approach of DREEM to propose a lightweight MT evaluation measure that employs only a raw corpus as an external resource. We adopt sentence similarity measures proposed by Song and Roth (2015) for a Semantic Textual Similarity (STS) task. They use word embeddings to align words so that the sentence similarity score takes near-synonymous expressions into account and propose three types of heuristics using m:n (average), 1:n (maximum) and 1:1 (Hungarian) alignments. It has been reported that sentence similarity calculated with a word alignment based on word embeddings shows high accuracy on STS tasks. 

We evaluated the word-alignment-based sentence similarity for MT evaluation to use the WMT12, WMT13, and WMT15 datasets of European--English translation and WAT2015 and NTCIR8 datasets of Japanese--English translation. Experimental results confirmed that the maximum alignment similarity outperforms previous word embeddings-based methods in European--English translation tasks and the average alignment similarity has the highest human correlation in Japanese--English translation tasks.
  
\section{Related Work}
Several studies have examined automatic evaluation of MT systems. The de facto standard automatic MT evaluation metrics BLEU \cite{Papineni:2002} may assign inappropriate score to a translation hypothesis that uses similar but different words because it considers only word n-gram precision \cite{Callison-Burch:2006}. METEOR-Universal \cite{Denkowski:2014} alleviates the problem of surface mismatch by using a thesaurus and a stemmer but it needs external resources, such as WordNet. In this work, we used a distributed word representation to evaluate semantic relatedness between the hypothesis and reference sentences. This approach has the advantage that it can be implemented only with only a raw monolingual corpus. \\
\indent To address the problem of word n-gram precision, \citet{wang-merlo:2016:HyTra6} propose to smooth it by word embeddings. They also employ maximum alignment between n-grams of hypothesis and reference sentences and a threshold to cut off n-gram embeddings with low similarity. Their work is similar to our maximum alignment similarity method, but they only experimented in European--English datasets, where maximum alignment works better than average alignment.\\
\indent The previous method most similar to ours is DREEM \cite{chen-guo:2015:ACL-IJCNLP}. It has shown to achieve state-of-the-art accuracy compared with popular metrics such as BLEU and METEOR. It uses various types of representations such as word and sentence representations. Word representations are trained with a neural network and sentence representations are trained with a recursive auto-encoder, respectively. DREEM uses cosine similarity between distributed representations of hypothesis and reference as a translation evaluation score. Both their and our methods employ word embeddings to compute sentence similarity score, but our method differs in the use of alignment and length penalty. As for alignment, we set a threshold to remove noisy alignments, whereas they use a hyper-parameter to down-weight overall sentence similarity. As for length penalty, we compared average, maximum, and Hungarian alignments to compensate for the difference between the lengths of translation hypothesis and reference, whereas they use an exponential penalty to normalize the length. \\
\indent Another way to improve the robustness of MT evaluation is to use a character-based model. CHRF \cite{popovic:2015:WMT} is one such metric that uses character n-grams. It is a harmonic mean of character n-gram precision and recall. It works well for morphologically rich languages. We, instead,  adopt a word-based approach because our target language, English, is morphologically simple but etymologically complex.

\section{Word-Alignment-Based Sentence Similarity using Word Embeddings}
\label{subsec:align}
In this section, we introduce word-alignment-based sentence similarity \cite{Song:2015} applied as an MT evaluation metrics.  Song and Roth \shortcite{Song:2015} propose to use word embeddings to align words in a pair of sentences. Their approach shows promising results in STS tasks. 

In MT evaluation, a word in the source language aligns to either a word or a phrase in the target language; therefore, it is not likely for a word to align with the whole sentence. Thus, we use several heuristics to constrain word alignment between the hypothesis and reference sentences.

In the following subsections, we present three sentence similarity measures. All of them use cosine similarity to calculate word similarity. To avoid alignment between unrelated words, we cut off word alignment whose similarity is less than a threshold value.

\subsection{Average Alignment Similarity}
\label{subsec:AAS}
First, the average alignment similarity (AAS) heuristic aligns a word with multiple words in a sentence pair. Similarity of words between a hypothesis sentence and a reference sentence is calculated. AAS is given by averaging word similarity scores of all combinations of words in $|x||y|$. 
\begin{equation}
	\mathrm{AAS}(x,y) = \frac{1}{|x||y|} \sum_{i=1}^{|x|} \sum_{j=1}^{|y|} \phi(x_i,y_j)
\end{equation}

\noindent Here, $x$ is a hypothesis and $y$ is a reference; and $x_{i}$ and $y_{j}$ represent words in each sentence.

\subsection{Maximum Alignment Similarity}
\label{subsec:MAS}
Second, we propose the maximum alignment similarity (MAS) heuristic averaging only the word that has the maximum similarity score of each aligned word pair. By definition, MAS itself is an asymmetric score so we symmetrize it by averaging the score in both directions. 
\begin{equation}
	\mathrm{MAS_{asym}}(a,b) = \frac{1}{|a|} \sum_{i=1}^{|a|} \max_j \phi(a_i,b_j)
\end{equation}
\begin{equation}
	\mathrm{MAS}(x,y) = \frac{1}{2} (\mathrm{MAS_{asym}}(x,y) + \mathrm{MAS_{asym}}(y,x))
\end{equation}
\noindent Here, $a$ and $b$ are words in a hypothesis and a reference sentence, respectively.

\begin{table*}[t]
 \begin{center}
  \begin{tabular}{|c|c|c|c|c|c|c|c|c|c|} \hline
    Evaluation Metrics & Fr-En & Fi-En & De-En & Cs-En & Ru-En & Average\\ \hline \hline
    Average Alignment Similarity      & 0.324 & 0.247 & 0.304 & 0.288 & 0.273 & 0.287\\ 
    Maximum Alignment Similarity & \bf{0.368} & \bf{0.355} & \bf{0.392} & 0.400 & \bf{0.349} & \bf{0.373} \\ 
   Hungarian Alignment Similarity & 0.223 & 0.211 & 0.259 & 0.251 & 0.239 & 0.237 \\ \hline
	BLEU \cite{stanojevic-EtAl:2015:WMT}                           & 0.358 & 0.308 & 0.360 & 0.391 & 0.329 & 0.349 \\ 
	DREEM \cite{chen-guo:2015:ACL-IJCNLP} & 0.362 & 0.340 & 0.368 & \bf{0.423} & 0.348 &  0.368\\ \hline
  \end{tabular}
 \end{center}
\begin{center}
\caption{Kendall's $\tau$ correlations of automatic evaluation metrics and official human judgements for the WMT15 dataset. (Fr: French, Fi: Finnish, De: German, Cs: Czech, Ru: Russian, En: English)}
\vspace*{-1em} 
\label{table:sentencewmtcorrelation}
\end{center}
\end{table*}

\begin{figure}[t]
 \includegraphics[width=8cm]{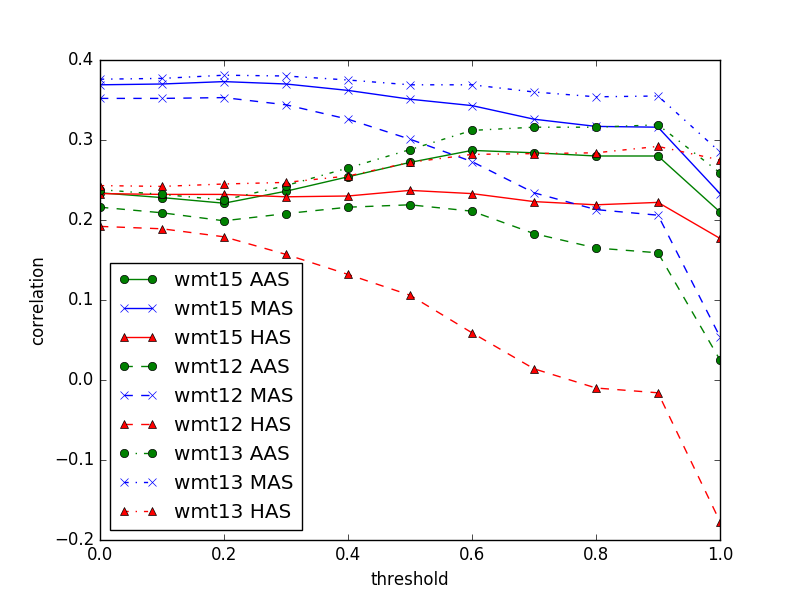}
  \caption{Correlation of each word-alignment-based method with varying the threshold for WMT datasets.}
 \label{fig:thresholdwmt}
\end{figure}

\begin{figure}[t]
 \includegraphics[width=8cm]{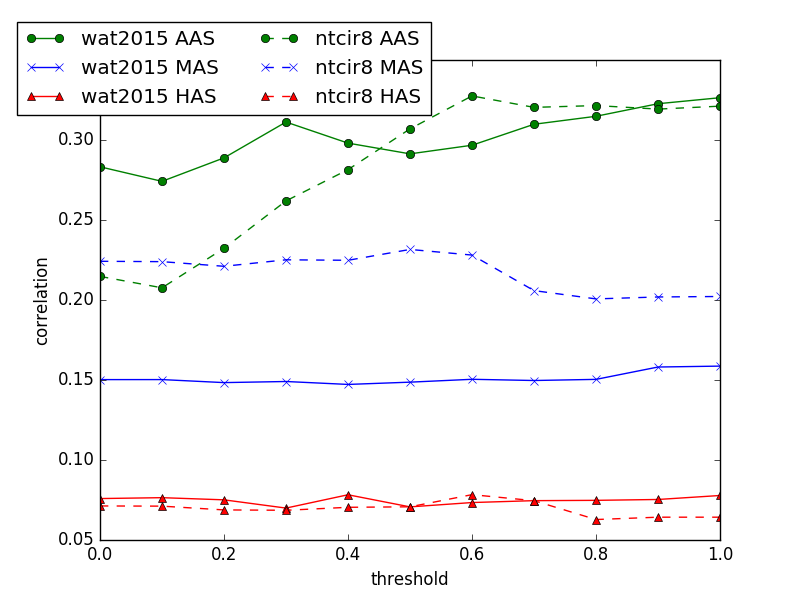}
  \caption{Correlation of each word-alignment-based method with varying the threshold for WAT2015 and NTCIR8 datasets.}
 \label{fig:thresholdacl}
\end{figure}

\subsection{Hungarian Alignment Similarity}
\label{subsec:HAS}
Third, we introduce the Hungarian alignment similarity (HAS) to restrict word alignment to 1:1. HAS formulates the task of word alignment as bipartite graph matching where  the words in a hypothesis and a reference are represented as nodes whose edges have weight $\phi(x_i,y_i)$. One-to-one word alignment is achieved by calculating maximum alignment of the perfect bipartite graph. For each word $x_i$ included in a hypothesis sentence, HAS chooses the word $h(x_i)$ in a reference sentence $y$ by the Hungarian method \cite{Kuhn:1955}. 
\begin{equation}
	\mathrm{HAS}(x,y) = \frac{1}{\min(|x|,|y|)} \sum_{i=1}^{|x|} \phi(x_i,h(x_i))
\end{equation}

\section{Experiment}
We report the results of MT evaluation in a European--English translation task of the WMT12, WMT13, and WMT15 datasets and Japanese--English task of WAT2015 and NTCIR8 datasets. For the WMT datasets, we compared our metrics with BLEU and DREEM taken from the official score of the WMT15 metric task \cite{stanojevic-EtAl:2015:WMT}. For WAT2015 and NTCIR8 datasets, the three types of proposed methods are compared.

\subsection{Experimental Setting}
\label{subsec:setting}
We used the WMT12, WMT13, and WMT15 datasets containing a total of 137,007 sentences in French, Finnish, German, Czech, and Russian translated to English. As Japanese--English translation datasets, WAT2015 includes 600 sentences and NTCIR8 includes 1,200 sentences. We measured correlation between human adequacy score and each of the evaluation metrics. We used Kendall's $\tau$ for segment-level evaluation. 
\begin{table*}[t]
 \begin{center}
  \begin{tabular}{|c|c|c|c|c|c|} \hline
    Evaluation Metrics & WMT12 & WMT13  & WMT15 & WAT2015 & NTCIR8\\ \hline \hline
    Average Alignment Similariy  & 0.211 & 0.312 & 0.287 & \bf{0.332} & \bf{0.343}\\ 
    Maximum Alignment Similarity & \bf{0.353} & \bf{0.381} & \bf{0.373} & 0.235 & 0.171\\ 
    Hungarian Alignment Similarity & 0.106 & 0.272 & 0.237 & 0.092 & 0.075 \\ \hline  
  \end{tabular}
 \end{center}
\begin{center}
\caption{Kendall's $\tau$ correlations of word-alignment-based methods and the official human judgements for each dataset. (WMT12, WMT13, and WMT15: European--English datasets, and WAT2015 and NTCIR8: Japanese--English datasets)}
\vspace*{-1em} 
\label{table:sentencewmt3correlation}
\end{center}
\end{table*}
We used a pre-trained model of word2vec using the Google News corpus for calculating word similarity using our proposed methods.\footnote{\url{https://code.google.com/archive/p/word2vec/}}

\subsection{Result}
\label{subsec:results}
Table \ref{table:sentencewmtcorrelation} shows a breakdown of correlation scores for each language pair in WMT15. MAS shows the best accuracy among all the proposed metrics for all language pairs. Its accuracy is better than that of DREEM for all language pairs except for Czech--English. This result shows that removal of noisy word embeddings by either using a threshold or 1:n alignment is important for European--English datasets.

Figure \ref{fig:thresholdwmt} shows correlation of word-alignment-based methods for WMT datasets with varying threshold values. For the WMT datasets, MAS has the highest correlation scores among the three word-alignment-based methods. A threshold value of 0.2 gives the maximum correlation for MAS for all WMT datasets.

Figure \ref{fig:thresholdacl} shows correlation of word-alignment-based methods for the two Japanese--English datasets with a varying threshold. Although MAS has the highest correlation for the WMT datasets, AAS has the highest correlation for the WAT2015 and NTCIR8 datasets.

Table \ref{table:sentencewmt3correlation} describes segment-level correlation results for WMT, WAT2015, and NTCIR8 datasets. MAS has the highest correlation score for the WMT datasets, whereas AAS has the highest correlation score for WAT2015 and NTCIR8 datasets.

\section{Discussion}
Figure \ref{fig:thresholdwmt} demonstrated that MAS and AAS are more stable than HAS for European--English datasets.  This may be because it is relatively easy for the AAS and MAS to perform word alignment using word embeddings in translation pairs of similar languages, but HAS suffers from alignment sparsity more than the other methods. In European--English translation, all the word-alignment-based methods perform poorly when using no word embeddings.

Unlike the European--English translation task, the Japanese--English translation task exhibits a different tendency. Figure \ref{fig:thresholdacl} shows the comparison between three types of word-alignment-based methods for each threshold. This is partly because word embeddings help evaluating lexically similar word pairs but fail to model syntactic variations. Also, we note that in Japanese--English datasets, AAS achieved the highest correlation. We suppose that this is because in Japanese--English translation, it is difficult to cover all the source information in the target language, resulting in misalignment of inadequate words by HAS and MAS. 

Table \ref{table:sentencewmt3correlation} shows that MAS performs stably on the WMT datasets. In particular, Kendall's $\tau$ score of HAS in WMT12 exhibits very low correlation. It seems that the 1:1 alignment is too strict to calculate sentence similarity in MT evaluation, while the 1:m (MAS) alignment performs well, possibly because of the removal of noisy word alignment. On the other hand, AAS is more stable than MAS and HAS for WAT2015 and NTCIR8 datasets. As a rule of thumb, AAS with high threshold values (0.6--0.9) shows stable high correlation across all language pairs, but if it is possible to use development data to tune the parameters, MAS with different values of thresholds should be considered.
 
\section{Conclusion}
In this paper, we presented word-alignment-based MT evaluation metrics using distributed word representations. In our experiments, MAS showed higher correlation with human evaluation than other automatic MT metrics such as BLEU and DREEM for European--English datasets. On the other hand, for Japanese--English datasets, AAS showed higher correlation with human evaluation than other metrics. These results indicate that appropriate word alignment using word embeddings is helpful in evaluating the MT output. 

\bibliography{acl2017}

\begin{thebibliography}{}
\expandafter\ifx\csname natexlab\endcsname\relax\def\natexlab#1{#1}\fi

\bibitem[{Callison-Burch et~al.(2006)Callison-Burch, Osborne, and
  Koehn}]{Callison-Burch:2006}
Chris Callison-Burch, Miles Osborne, and Philipp Koehn. 2006.
\newblock {Re-evaluating the Role of BLEU in Machine Translation Research}.
\newblock In {\em Proceedings of the 11th Conference of the European Chapter of
  the Association for Computational Linguistics\/}. pages 249--256.

\bibitem[{Chen and Guo(2015)}]{chen-guo:2015:ACL-IJCNLP}
Boxing Chen and Hongyu Guo. 2015.
\newblock {Representation Based Translation Evaluation Metrics}.
\newblock In {\em Proceedings of the 53rd Annual Meeting of the Association for
  Computational Linguistics and the 7th International Joint Conference on
  Natural Language Processing (Volume 2: Short Papers)\/}. pages 150--155.

\bibitem[{Denkowski and Lavie(2014)}]{Denkowski:2014}
Michael Denkowski and Alon Lavie. 2014.
\newblock {Meteor Universal: Language Specific Translation Evaluation for Any
  Target Language}.
\newblock In {\em Proceedings of the Ninth Workshop on Statistical Machine
  Translation\/}. pages 376--380.

\bibitem[{Kuhn(1955)}]{Kuhn:1955}
Harold~W. Kuhn. 1955.
\newblock {The Hungarian Method for the Assignment Problem}.
\newblock In {\em Naval Research Logistics Quarterly\/}. pages 83--97.

\bibitem[{Papineni et~al.(2002)Papineni, Roukos, Ward, and Zhu}]{Papineni:2002}
Kishore Papineni, Salim Roukos, Todd Ward, and Wei-Jing Zhu. 2002.
\newblock {BLEU: a Method for Automatic Evaluation of Machine Translation}.
\newblock In {\em Proceedings of the 40th annual meeting on association for
  computational linguistics. Association for Computational Linguistics\/}.
  pages 311--318.

\bibitem[{Popovi\'{c}(2015)}]{popovic:2015:WMT}
Maja Popovi\'{c}. 2015.
\newblock {ChrF: Character n-gram F-score for Automatic MT Evaluation}.
\newblock In {\em Proceedings of the Tenth Workshop on Statistical Machine
  Translation\/}. pages 392--395.

\bibitem[{Song and Roth(2015)}]{Song:2015}
Yangqui Song and Dan Roth. 2015.
\newblock {Unsupervised Sparse Vector Densification for Short Text Similarity}.
\newblock In {\em Proceedings of the 2015 Annual Conference of the North
  American Chapter of the ACL\/}. pages 1275--1280.

\bibitem[{Stanojevi\'{c} et~al.(2015)Stanojevi\'{c}, Kamran, Koehn, and
  Bojar}]{stanojevic-EtAl:2015:WMT}
Milo\v{s} Stanojevi\'{c}, Amir Kamran, Philipp Koehn, and Ond\v{r}ej Bojar.
  2015.
\newblock {Results of the WMT15 Metrics Shared Task}.
\newblock In {\em Proceedings of the Tenth Workshop on Statistical Machine
  Translation\/}. pages 256--273.

\bibitem[{Wang and Merlo(2016)}]{wang-merlo:2016:HyTra6}
Haozhou Wang and Paola Merlo. 2016.
\newblock {Modifications of Machine Translation Evaluation Metrics by Using
  Word Embeddings}.
\newblock In {\em Proceedings of the Sixth Workshop on Hybrid Approaches to
  Translation (HyTra6)\/}. pages 33--41.

\end{thebibliography}
\bibliographystyle{acl_natbib}

\appendix

%
%
%
%
%
\end{document}